\documentclass[10pt,twocolumn,letterpaper]{article}
\usepackage{cvpr}

\usepackage[utf8]{inputenc} 
\usepackage[T1]{fontenc}    
\usepackage{url}            
\usepackage{booktabs}       
\usepackage{amsfonts}       
\usepackage{nicefrac}       
\usepackage{microtype}      
\usepackage{xcolor}
\usepackage{bbm}
\usepackage{amsmath}
\usepackage{xspace}
\usepackage{graphicx}
\usepackage{bm}
\usepackage{xspace}
\usepackage{multicol}
\usepackage{multirow}
\usepackage[font=small]{caption} 
\newcommand{\CloseNei}{\mathbf{C}_i}
\newcommand{\BackNei}{\mathbf{B}_i}
\newcommand{\VI}{\mathbf{v}_i}
\newcommand{\BmTheta}{\bm{\theta}}
\newcommand{\EI}{\mathbf{e}_i}

\newcommand{\ShortName}{VIE\xspace}

\usepackage[breaklinks=true,bookmarks=false]{hyperref}
\cvprfinalcopy 
\def\cvprPaperID{8837} 

\ifcvprfinal\fi
\title{Unsupervised Learning from Video with Deep Neural Embeddings}

\author{%
Chengxu Zhuang$^1$ \quad 
Tianwei She$^1$  \quad 
Alex Andonian$^2$  \quad 
Max Sobol Mark$^1$  \quad 
Daniel Yamins$^1$\\
{$^1$Stanford University \quad $^2$ MIT}\\
{\tt\small \{chengxuz, shetw, joelmax, yamins\}@stanford.edu  \quad andonian@mit.edu}
}

\begin{document}

\maketitle
\thispagestyle{empty}

\begin{abstract}
Because of the rich dynamical structure of videos and their ubiquity in everyday life, it is a natural idea that video data could serve as a powerful unsupervised learning signal for visual representations.
However, instantiating this idea, especially at large scale, has remained a significant artificial intelligence challenge. 
Here we present the Video Instance Embedding (VIE) framework, which trains deep nonlinear embeddings on video sequence inputs.
By learning embedding dimensions that identify and group similar videos together, while pushing inherently different videos apart in the embedding space, VIE captures the strong statistical structure inherent in videos, without the need for external annotation labels.
We find that, when trained on a large-scale video dataset, VIE yields powerful representations both for action recognition and single-frame object categorization, showing substantially improving on the state of the art wherever direct comparisons are possible. 
We show that a two-pathway model with both static and dynamic processing pathways is optimal, provide analyses indicating how the model works, and perform ablation studies showing the importance of key architecture and loss function choices.
Our results suggest that deep neural embeddings are a promising approach to unsupervised video learning for a wide variety of task domains.
\end{abstract}

\section{Introduction} \label{sec:intro}
A video's temporal sequence often contains information about dynamics and events in the world that is richer than that in its unordered set of frames. 
For example, as objects and agents move and interact with each other, they give rise to characteristic patterns of visual change that strongly correlate with their visual and physical identities, including object category, geometric shape, texture, mass, deformability, motion tendencies, and many other properties.
It is thus an attractive hypothesis that ubiquitiously available natural videos could serve as a powerful signal for unsupervised learning of visual representations for both static and dynamic visual tasks.

However, it has been challenging to embody this hypothesis in a concrete neural network that can consume unlabelled video data to learn useful feature representations, especially in the context of at-scale real-world applications.
 Perhaps the biggest source of difficulty in making progress on unsupervised video learning, though, is that unsupervised learning has presented a formidable challenge even for the case of single static images.  
Even for the case of single static images, the gap in representational power between the features learned by unsupervised and supervised neural networks has been very substantial, to the point where the former were unsuitable for use in any at-scale visual task. 
However, recent advances in learning with deep visual embeddings have begun to produce unsupervised representations that rival the visual task transfer power of representations learned by their supervised counterparts~\cite{wu2018unsupervised,wu2018improving,zhuang2019local,caron2018deep}.
These methods leverage simple but apparently strong heuristics about data separation and clustering to iteratively bootstrap feature representations that increasingly better capture subtle natural image statistics.  
As a result, it is now possible to obtain unsupervised deep convolutional neural networks that outperform ``early modern'' deep networks (such as AlexNet in~\cite{hinton2012deep}) on challenging recognition tasks such as ImageNet, even when the latter are trained in a supervised fashion.
Moreover, works in supervised video classification, action recognition and video captioning have proposed novel combinations of two- and three-dimensional convolutional structures that are increasingly well-suited to video feature extraction~\cite{slowfast}. 

In this work, we show how the idea of deep unsupervised embeddings can be used to learn features from videos, introducing the Video Instance Embedding (VIE) framework.  
In VIE, videos are projected into a compact latent space via a deep neural network, whose parameters are then tuned to optimally distribute embedded video instances so that similar videos aggregate while dissimilar videos separate. 
We find that VIE learns powerful representations for transfer learning to action recognition in the large-scale Kinetics dataset, as well as for single-frame object classification in the ImageNet dataset. 
Moreover, where direct comparison to previous methods is possible, we find that VIE substantially improves on the state-of-the-art. 
We evaluate several possibilities for the unsupervised VIE loss function, finding that those that have been shown to be most effective in the single-frame unsupervised learning~\cite{zhuang2019local} are also most effective in the video context.
We also explore several neural network embedding and frame sampling architectures, finding that different temporal sampling statistics are better priors for different transfer tasks, and that a two-pathway static-dynamic architecture is optimal. 
Finally, we present analyses of the learned representations giving some intuition as to how the models work, and a series of ablation studies illustrating the importance of key architectural choices.
Codes can be found at \url{https://github.com/neuroailab/VIE}.

\section{Related Work} \label{sec:relat}
\textbf{Unsupervised Learning of Deep Visual Embeddings.}
In this work, we employ a framework derived from ideas first introduced in the recent literature on unsupervised learning of embeddings for images~\cite{wu2018unsupervised}. 
In the Instance Recognition (IR) task, a deep nonlinear image embedding is trained to maximize the distances between different images while minimizing distances between augmentations (e.g. crops) of a given image, thus maximizing the network's ability to recognize individual image instances.
In the Local Aggregation (LA) task~\cite{zhuang2019local}, the embedding loss also allows selected groups of images to aggregate, dynamically determining the groupings based on a local clustering measure.
Conceptually, the LA approach resembles a blending of IR and the also-recent DeepCluster method~\cite{caron2018deep}, and is more powerful than either IR or DeepCluster alone, achieving state-of-the-art results on unsupervised learning with images.
Another task Contrastive Multiview Coding (CMC)~\cite{CMC} achieves similar performance to LA within this embedding framework while aggregating different views of a given image.
CMC have also been directly applied to videos, where single frame is clustered with its future frame and its corresponding optical flow image.
The VIE framework allows the use of any of these embedding objectives, and we test several of them here. 
Other recently proposed methods such as~\cite{henaff2019data} and~\cite{bachman2019learning} have achieved comparable results to LA and CMC through optimizing the mutual information of different views of the images, though they use much deeper and more complex architectures.

\textbf{Supervised Training of Video Networks.}
Neural networks have been used for a variety of supervised video tasks, including captioning~\cite{krishna2017dense} and 3D shape extraction~\cite{matsuyama2004real,akbarzadeh2006towards}, but the architectures deployed in those works are quite different from those used here.
The structures we employ are more directly inspired by work on supervised action recognition.  
A core architectural choice explored in this literature is how and where to use 2D single-frame vs 3D multi-frame convolution. 
A purely 2D approach is the Temporal Relational Network (TRN)~\cite{TRN}, which processes aggregates of 2D convolutional features using MLP readouts.
Methods such as I3D~\cite{carreira2017quo} have shown that combinations of both 2D and 3D can be useful, deploying 2D processing on RGB videos and 3D convolution on an optical flow component. 
A current high-performing architecture for action recognition is the SlowFast network~\cite{slowfast}, which computes mixed 2D-3D convolutional features from sequences of images at multiple time scales, including a slow branch for low frequency events and a fast branch for higher-frequency events.
The dynamic branch of our two-pathway architecture is chosen to mimic the most successful SlowFast network. 
However, we find it useful to include in our architecture a static pathway that is not equivalent to either of the SlowFast branches.

\textbf{Unsupervised Learning on Videos.}
The literature on unsupervised video learning is too extensive to review comprehensively here, so we focus our discussion on several of the most relevant approaches. 
Temporal autoencoders such as PredNet~\cite{lotter2016deep}, PredRNN~\cite{wang2017predrnn}, and PredRNN++~\cite{wang2018predrnn++} are intriguing but have not yet evidenced substantial transfer learning performance at scale. 
Transfer learning results have been generated from a variety of approaches including the Geometry-Guided CNN~\cite{gan2018geometry}, motion masks~\cite{pathak2017learning}, VideoGAN~\cite{vondrick2016generating}, a pairwise-frame siamese triplet network~\cite{wang2015unsupervised}, the Shuffle and Learn approach~\cite{misra2016shuffle}, and the Order Prediction Network (OPN)~\cite{lee2017unsupervised}. 
More recent works, including the Video Rotations Prediction task (3DRotNet)~\cite{Jing2018}, the Video Motion and Appearance task (MoAp)~\cite{Wang2019}, the Space-Time Puzzle task (ST-puzzle)~\cite{Kim2018}, and the Dense Predictive Coding (DPC) task~\cite{DPC}, have reported improved performance, with the help of pretraining on large-scale datasets and using spatiotemporal network architectures.
All these works only operate on relationships defined within a single video, differentiating them from \ShortName, which exploits the relationships both within and between different videos through a loss function defined on the \emph{distribution} of video embeddings.

\section{Methods} \label{sec:method}

\begin{figure*}
\begin{center}
\includegraphics[width=.8\textwidth] {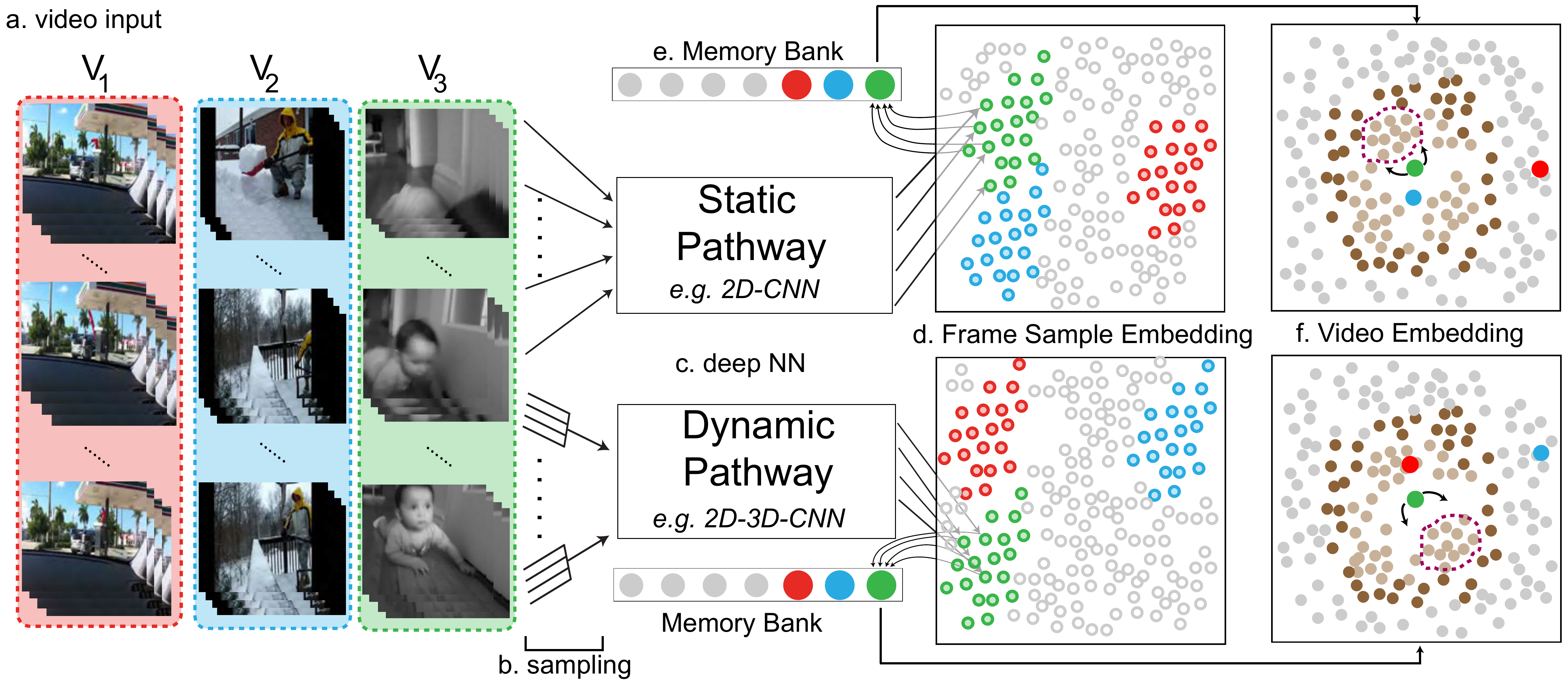}
\end{center}
\caption{\textbf{Schematic of the Video Instance Embedding (VIE) Framework.} \textbf{a.} Frames from individual videos ($\mathbf{v}_1$, $\mathbf{v}_2$, $\mathbf{v}_3$) are \textbf{b.} sampled into sequences of varying lengths and temporal densities, and input into \textbf{c.} deep neural network pathways that are either static (single image) or dynamic (multi-image).
\textbf{d.} Outputs of frame samples from either pathway are vectors in the $D$-dimensional unit sphere $S^D \subset \mathbf{R}^{D+1}$. 
The running mean value of embedding vectors are calculated over online samples for each video, \textbf{e.} stored in a memory bank, and \textbf{f.} at each time step compared via unsupervised loss functions in the video embedding space. 
The loss functions require the computation of distribution properties of embedding vectors. 
For example, the Local Aggregation (LA) loss function involves the identification of Close Neighbors $\CloseNei$ (light brown points) and Background Neighbors $\BackNei$ (dark brown points), which are used to determine how to move target point (green) relative to other points (red/blue). 
}
\label{fig:schema}
\vspace{-4mm}
\end{figure*}

\textbf{VIE Embedding framework.} 
The general problem of unsupervised learning from videos can be formulated as learning a parameterized function $\phi_{\BmTheta}(\cdot)$ from input videos $\mathbf{V}=\{\VI | i=1, 2, .., N\}$, where each $\VI$ consists of a sequence of frames $\{f_{i,1}, f_{i,2}, ..., f_{i,m_i}\}$.
Our overall approach seeks to embed the videos $\{\VI\}$ as feature vectors $\mathbf{E}=\{\EI\}$ in the $D$-dimension unit sphere $S^{D} = \{x \in \mathbf{R}^{D+1} \text{ with } ||x||^2_2 = 1 \}$.
This embedding is realized by a neural network $\phi_{\BmTheta}: \VI \mapsto S^D$ with weight parameters $\BmTheta$, that receives a sequence of frames $\mathbf{f}=\{f_1, f_2, ..., f_L\}$ and outputs $e=\phi_{\BmTheta}(\mathbf{f}) \in S^D$.
Although the number of frames in one video can be arbitrary and potentially large, $L$ must usually be fixed for most deep neural networks.
Therefore, the input $\mathbf{f}$ on any single inference pass is restricted to a subset of frames in $\mathbf{v}$ chosen according to a \emph{frame sampling strategy} $\rho$ --- that is, a random variable function such that $\mathbf{v}' \subset \mathbf{v}$ for all samples $\mathbf{v}'$ drawn from $\rho(\mathbf{v})$.
Given $\rho$, we then define the associated \emph{\textbf{Video Instance Embedding}} (VIE) $\mathbf{e}$ for video $\mathbf{v}$ as the normed (vector-valued) expectation of $e$ under $\rho$, i.e.
\begin{equation} \label{equ:vie}
\mathbf{e} = \frac{\mathrm{E}_\rho
[\phi_{\BmTheta}(\mathbf{f})]}{||\mathrm{E}_\rho [\phi_{\BmTheta}(\mathbf{f})]||_2} \in S^D.
\end{equation}
In addition to choosing $\phi_{\BmTheta}$ and $\rho$, we must choose a loss function $\mathcal{L}: \mathbf{E} \mapsto \mathbf{R}$ such optimizing $\mathcal{L}$ with respect to $\BmTheta$ will cause statistically related videos to be grouped together and unrelated videos to be separated.
Note that in theory this function depends on all of embedded vectors, although in practice it is only ever evaluated on a stochastically-chosen batch at any one time, with dataset-wide effects captured via a memory bank.

In the following subsections, we describe natural options for these main components (architecture $\phi_{\BmTheta}$, sampling strategy $\rho$, and loss function $\mathcal{L}$).
As shown in Section~\ref{sec:results}, such choices can not only influence the quality of learned representations, but also change the focus of the representations along the spectrum between static and dynamic information extraction.

\textbf{Architecture $\phi$ and sampling strategy $\rho$.}
Recent exploration in supervised action recognition has provided a variety of choices for $\phi_{\BmTheta}$.
Although very complex network options are possible~\cite{carreira2017quo}, since this work is an initial exploration of the interplay between video processing architecture and unsupervised loss functions, we have chosen to concentrate on five simple but distinct model families. 
They are differentiated mainly by how their frame sampling assumptions represent different types of temporal information about the inputs:

1. \emph{Single-frame 2D-CNNs.}
Although deep 2D convolutional neural networks (CNNs) taking one frame as input ignore temporal information in the videos, they can still leverage context information and have achieved nontrivial performance on action recognition datasets~\cite{carreira2017quo}.
They are also useful baselines for measuring the effect of including temporal information.

2. \emph{3D-CNNs with dense equal sampling.}
3D-CNNs, with spatiotemporal filters, can be applied to dense evenly-sampled frames to capture fine-grained temporal information.
This architecture has proven useful in the R3D networks of~\cite{tran2018closer} and the 3DResNets of~\cite{Hara2018}.

3. \emph{Shared 2D-CNNs with sparse unequal sampling.}
The Temporal Relation Network bins videos into half-second clips, chooses $L$ consecutive bins, and randomly samples one frame from each bin.
A shared 2D-CNN is then applied to each frame and its outputs are concatenated in order and fed into an MLP that creates the embedding.
Unlike the dense sampling approach, this method can capture long-range temporal information through sparse sampling, but as the intervals between frames are uneven, the temporal signal can be noisy.

4. \emph{2D-3D-CNNs with sparse equal sampling.}
Addressing the issue of noisy temporal information in the third family, the Slow branch of the SlowFast~\cite{slowfast} architecture samples frames equally but sparsely from the input video.
These frames are then passed through 2D-CNNs with spatial pooling, applying 3D convolutions downstream once spatial redundancy is reduced. 

5. \emph{Multi-pathway architectures.} 
Combination architectures allow the exploitation of the multiple temporal scales present in natural videos, including the SlowFast approach~\cite{slowfast} (combining 2 and 4), and true two-pathway networks with both static and dynamic pathways (combining 1, 2, and 4). 

In our experiments (\S\ref{sec:results}), we implement these models with CNN backbones that, while accommodating small unavoidable differences due to the input structure, are otherwise as similar as possible, so that the qualities of learned representations can be fairly compared.

\textbf{Loss function $\mathcal{L}$.}
Recent work in unsupervised learning with single images has found useful generic metrics for measuring the quality of deep visual embeddings~\cite{wu2018unsupervised, zhuang2019local}, including the Instance Recognition (IR) and Local Aggegation (LA) loss functions.
These methods seek to group similar inputs together in the embedding space, while separating dissimilar inputs.
They are based on a simple probabilistic perspective for interpreting compact embedding spaces~\cite{wu2018unsupervised, zhuang2019local}.
Specifically, the probability that an arbitrary feature $e$ is recognized as a sample of $\VI$ is defined to be: 
\begin{equation}
    P(i | e, \mathbf{E}) = \frac{\mathrm{exp}(\EI^T e / \tau)}{\sum_{j=1}^N\mathrm{exp}(\mathbf{e}_j^T e / \tau)}
    \label{equ:single}
\end{equation}
where temperature $\tau \in (0, 1]$ is a fixed scale hyperparameter.
Both $\{\EI\}$ and $e$ are projected onto the unit sphere $S^D$.
With this definition in mind, we can define the IR and LA loss functions, adapted to the video context via eq. \ref{equ:vie}.

\emph{IR algorithm.} The VIE-version of the IR loss is:
\begin{equation} \label{equ:IR_loss}
\mathcal{L}^{\mathrm{IR}}(\VI, \mathbf{E}) = -\log P(i|e, \mathbf{E}) + \lambda \|\BmTheta\|^2_2
\end{equation}
where $\lambda$ is a regularization hyperparameter, and where for computational efficiency the denominator in $P(i|e,\mathbf{E})$ is estimated through randomly choosing a subset of $Q$ out of all $N$ terms (see~\cite{wu2018unsupervised} for further details).
Intuitively, optimizing this loss will group embeddings of frame groups sampled from the same video, which then implicitly gathers other similar videos.

\emph{LA algorithm.} Local Aggregation augments IR by allowing for a more flexible dynamic detection of which datapoints should be grouped together.
Define the probability that a feature $e$ is recognized as being in a set of videos $\mathbf{A}$ as $P(\mathbf{A} | e, \mathbf{E}) = \sum_{i \in \mathbf{A}} P(i | e, \mathbf{E})$.
For a video $\VI$ and its embedding $\EI$, the LA algorithm identifies two sets of neighbors, the \emph{close neighbors} $\CloseNei$ and \emph{background neighbors} $\BackNei$.
$\CloseNei$ is computed via dynamic online $k$-means clustering, and identifies datapoints expected to be ``especially similar'' to $\VI$; $\BackNei$ is computed via $k$-nearest neighbors method, and sets the scale of distance in terms of which closeness judgements are measured. 
Given these two neighbor sets, the local aggregation loss function measures the negative log-likelihood of a point being recognized as a close neighbor given that is already a background neighbor:
\begin{equation}\label{equ:LA_loss}
\mathcal{L}^{\mathrm{LA}}(x_i, \mathbf{E}) = - \log \frac{P(\CloseNei \cap \BackNei | \VI, \mathbf{E})}{P(\BackNei | \VI, \mathbf{E})} + \lambda \|\BmTheta\|^2_2.
\end{equation}
Intuitively, the LA loss function encourages the emergence of soft clusters of datapoints at multiple scales.  
See~\cite{zhuang2019local} for more details on the LA procedure.

\textbf{Memory Bank.}
Both the IR and LA loss functions implicitly require access to all the embedded vectors $\mathbf{E}$ to calculate their denominators.
However, recomputing $\mathbf{E}$ is intractable for big dataset. 
This issue is addressed, as in~\cite{wu2018unsupervised, wu2018improving, zhuang2019local}, by approximating $\mathbf{E}$ with a memory bank $\bar{\mathbf{E}}$ keeping a running average of the embeddings.

\section{Experiments and Results} \label{sec:results}

\textbf{Experimental settings.}
To train our models, we use the Kinetics-400 dataset~\cite{kinetics}, which contains approximately 240K training and 20K validation videos, each around 10 seconds in length and labeled in one of 400 action categories.
After downloading, we standardize videos to a framerate of 25fps and reshape all frames so that the shortest edge is 320px.
After sampling according to the frame sampling strategy for each model architecture, we then apply the spatial random cropping and resizing method used in~\cite{slowfast}. 
Following~\cite{zhuang2019local} we also apply color noise and random horizontal flip, using the same spatial window and color noise parameters for all the frames within one video.
At test time, we sample five equally-spaced sequences of frames, resize them so that their shortest side is 256px, and take center $224 \times 224$ crop.
Softmax logits for the five samples are then averaged to generate the final output prediction.
 
We use ResNet-18v2~\cite{he2016identity} as the convolutional backbone for all our model families, to achieve a balance between model performance and computation efficiency. 
Implementations for the different model families are denoted \ShortName-Single (``Family 1''), \ShortName-3DResNet\footnote{To be comparable to previous work, our 3DResNet uses a lower input resolution, $112 \times 112$. See the supplementary material.} (``Family 2''), \ShortName-TRN (``Family 3''), \ShortName-Slow (``Family 4''), and \ShortName-SlowFast (``Family 5'').
Two-pathway models are created by concatenating single- and multi-frame network outputs, yielding \ShortName-TwoPathway-S (combining \ShortName-Single and \ShortName-Slow) and \ShortName-TwoPathway-SF (combining \ShortName-Single and \ShortName-Slowfast).
We follow~\cite{zhuang2019local} for general network training hyperparameters including initial learning rate, optimizer setting, learning rate decay schedule, batch size, and weight decay coefficient.
Further details of model architecture and training can be found in the supplementary material.

\textbf{Transfer to action recognition on Kinetics.}
After training all models on the unlabelled videos from Kinetics, we evaluate the learned representations by assessing transfer learning performance to the Kinetics action recognition task.
To compare our method to previous work on this task, we reimplement three strong unsupervised learning algorithms: OPN~\cite{lee2017unsupervised}, 3DRotNet~\cite{Jing2018}, and RotNet~\cite{rotnet}.
As OPN and RotNet require single-frame models, we use ResNet18 as their visual backbones.
For 3DRotNet, we use 3DResNet18 with the same input resolution as \ShortName-3DResNet.
Reimplementation details are in the supplementary material.
Transfer learning is assessed through the standard transfer procedure of fixing the learned weights and then training linear-softmax readouts from different layers of the fixed model.
We implemented this procedure following the choices in ~\cite{zhuang2019local}, but using the Kinetics-specific data augmentation procedure described above. 
Since some of the models generate outputs with a temporal dimension, directly adding a fully-connected readout layer would lead to more trainable readout parameters as compared to single frame models.
To ensure fair comparisons, we thus average the features of such models along the temporal dimension before readout.

\setlength{\fboxrule}{1pt}%
\begin{table*}[t]
\centering
\begin{tabular}{c|c|ccc|ccc}
\hline
Datasets & \multicolumn{4}{|c|}{Kinetics} & \multicolumn{3}{c}{ImageNet} \\
\hline
Metric & Super. & Conv3 & Conv4 & Conv5 & Conv3 & Conv4 & Conv5 \\
\hline\hline
Random-Single  & -- & 9.40 & 8.43 & 6.84 & 7.98 & 7.78 & 6.23 \\
OPN-Single*~\cite{lee2017unsupervised} & -- & 16.84 & 20.82 & 20.86 & 13.01 & 17.63 & 18.29 \\
RotNet-Single*~\cite{rotnet}  & -- & 26.25 & 30.27 & 23.33 & 25.77 & 27.59 & 16.13 \\
3DRotNet-3DResNet*~\cite{Jing2018} & -- & 28.30 & 29.33 & 19.33 & 23.34 & 22.05 & 12.45 \\
\hline
\ShortName-Single (IR) & 57.59 & 23.50 & 38.72 & 43.85 & 22.85 & 40.49 & 40.43 \\
\ShortName-Single & 57.59 & 23.84 & 38.25 & 44.41 & 25.02 & 40.49 & 42.33 \\
\ShortName-TRN & 59.43 & 25.72 & 39.38 & 44.91 & \textbf{27.24} & 40.28 & 37.44 \\
\ShortName-3DResNet & 53.22 & \textbf{33.01} & 41.34 & 43.40 & 30.18 & 35.37 & 32.62 \\
\ShortName-Slow & 60.84 & 24.80 & 40.48 & 46.36 & 20.10 & 37.02 & 37.45 \\
\ShortName-Slowfast & \fboxsep=1pt\fcolorbox{red}{white}{\textbf{62.36}} & 28.68 & 42.07 & 47.37 & 22.61 & 36.84 & 36.60 \\
\ShortName-TwoPathway-S & -- & 26.38 & 41.80 & 47.13 & 23.98 & 40.52 &  \fboxsep=1pt\fcolorbox{red}{white}{\textbf{44.02}} \\
\ShortName-TwoPathway-SF & -- & 29.89 & \textbf{43.50} &  \fboxsep=1pt\fcolorbox{red}{white}{\textbf{48.53}} & 23.23 & \textbf{40.73} & 43.69 \\
\hline
Supervised-Single & \multicolumn{4}{|c|} {\multirow{5}{*}{--}} & 22.32 & 37.82 & 38.26 \\
Supervised-TRN & \multicolumn{4}{|c|}{} & 22.82 & 41.13 & 39.15 \\
Supervised-3DResNet & \multicolumn{4}{|c|}{} & 28.09 & 34.40 & 30.56 \\
Supervised-Slow & \multicolumn{4}{|c|}{} & 21.86 & 40.77 & 32.87 \\
Supervised-SlowFast & \multicolumn{4}{|c|}{} & 20.25 & 37.41 & 30.75 \\
\hline
\end{tabular}
\caption{
Top-1 transfer learning accuracy (\%) on the Kinetics and ImageNet validation sets.
``Random'' means a randomly initialized ResNet-18 without any training.
``Supervised-*'' means trained on Kinetics for action recognition.
Our supervised performance is not directly comparable to~\cite{slowfast} due to different visual backbones used.
*: These numbers are generated by us.
}
\label{tab:transKN}
\end{table*}

Results are shown in Table~\ref{tab:transKN}. 
All \ShortName variants show significantly better performance than OPN, RotNet, and 3DRotNet.
Multi-frame models substantially outperform single-frame versions, an improvement that cannot be explained by the mere presence of additional frames (see the supplementary material Table S1).
The two-pathway models achieve the highest performance, with a maximum accuracy of approximately 48.5\%.
The rank order of unsupervised performances across VIE variants are aligned with that of supervised counterparts, indicating that the unsupervised training procedure takes advantage of increased architectural power when available. 
The LA-based \ShortName-Single model performs better than the IR-based model, consistent with the gap on static object recognition~\cite{zhuang2019local}.
Finally, though previous work on unsupervised video learning has started to leverage large-scale datasets such as Kinetics~\cite{DPC, Jing2018, Kim2018}, the transfer learning performance of the trained models to Kinetics action recognition has never been reported. We therefore hope these results are useful both for understanding the effect of architecture on representation quality and providing a strong unsupervised benchmark for future works.

\textbf{Transfer to action recognition on UCF101 and HMDB51.}
To compare \ShortName to previous methods, we evaluate results on the more commonly-used UCF101~\cite{ucf101} and HMDB51~\cite{HMDB51} action recognition benchmarks.
We initialize networks by pretrained weights on Kinetics and then finetune them on these datasets, following the procedures used for the most recent work~\cite{Jing2018, Wang2019, Kim2018}.
We notice that details of the data augmentation pipelines used during finetuning can influence final results. 
Most importantly, we find having color-noise augmentation can improve the finetuing performance. 
However, the augmentation techniques have not been carefully controlled in previous works. For example, ST-puzzle~\cite{Kim2018}, 3DRotNet~\cite{Jing2018}, and MoAp~\cite{Wang2019} only use the usual random crop and horizontal flip augmentations, while DPC~\cite{DPC} also uses color-noise. 
To ensure that our comparisons to these algorithms are fair, we therefore test our models with both augmentation pipelines.
Details can be found in the supplement.

Table~\ref{tab:finetuneUH_usual} and~\ref{tab:finetune_color} show that \ShortName substantially outperforms other methods.
Making these comparisons requires some care, because results reported in previous works are often confounded by the variation of multiple factors at once, making it hard to determine if improvements are really due to a better algorithm, rather than a larger training dataset, a more powerful network architecture, or a difference in input data types.
First, holding network architecture and training dataset fixed, 
\ShortName-3DResNet surpasses the previous state-of-the-art (ST-Puzzle) by \textbf{6.5\%} on UCF101 and \textbf{11.1\%} on HMDB51, approaching the supervised upper-bound. 
With the better augmentation pipeline, the improvement on UCF101 is \textbf{9.7\%}.
3DRotNets are trained on Kinetics-600, which contains more than twice as much training data and 50\% more categories than that used for ST-Puzzle and VIE models, and are trained with larger inputs (64-frame RGB vs VIE's 16-frame inputs).
Nonetheless, \ShortName still shows improvement of more than \textbf{6.2\%} when compared with comparable input type.
3DRotNet also reports results for a larger fused model trained with both 64-frame RGB and frame-difference inputs. \ShortName-Slowfast nonetheless substantially outperforms this model, using fewer trainable parameters (21M vs 32M) and much less training data (Kinetics-400 vs Kinetics-600).
When compared to the very recent DPC algorithm~\cite{DPC} in Table~\ref{tab:finetune_color}, \ShortName shows improvement of more than \textbf{7.3\%}, even though DPC uses more frames (40 v.s. 16) as inputs and adds an additional recurrent cell between the action readout layer and its 3DResNet.
Again, \ShortName makes good use of more complex architectures, as can be seen in the SlowFast vs ResNet18 comparison. 
Moreover, TwoPathway-SF achieves even better performance than SlowFast.

\setlength{\fboxrule}{1pt}%
\begin{table}[t]
\centering
\begin{tabular}{c|c|cc}
\hline
Networks & Algorithms & UCF & HMDB \\
\hline
\multirow{2}{*}{AlexNet$^{\dag}$} & CMC~\cite{CMC} & 59.1 & 26.7 \\
 & OPN~\cite{lee2017unsupervised} & 56.3 & 22.1 \\
\hline
 \multirow{2}{*}{C3D} & MoAp~\cite{Wang2019} & 61.2 & 33.4 \\
& ST-puzzle~\cite{Kim2018} & 60.6 & 28.3 \\
\hline
\multirow{6}{*}{3DResNet} & Scratch & 47.4 & 21.5 \\
 & ST-puzzle & 65.8 & 33.7 \\
 & 3DRotNet~\cite{Jing2018} & 62.9 & 33.7 \\
 & 3DRotNet(64f)~\cite{Jing2018} & 66.0 & 37.1 \\
 & VIE (ours) & \textbf{72.3} & \textbf{44.8} \\
 & \emph{Supervised} & \emph{84.4} & \emph{58.7} \\
\hline
\multirow{3}{*}{SlowFast} & Scratch & 55.8 & 21.4 \\
 & VIE (ours) & \textbf{77.0} & \textbf{46.5} \\
 & \emph{Supervised} & \emph{88.4} & \emph{68.4} \\
\hline
\multirow{3}{*}{ResNet18} & Scratch & 46.7 & 17.3 \\
 & VIE (ours) & 71.2 & 38.4 \\
 & \emph{Supervised} & \emph{81.0} & \emph{49.9} \\
\hline
\multicolumn{2}{c|}{VIE-TwoPathway-SF} & \textbf{78.2} & \textbf{50.5} \\
\hline
\end{tabular}
\caption{
Top-1 finetuning results on UCF101 and HMDB51 datasets using models pretrained on Kinetics \textbf{without} color-noise augmentation.
We also provide performance of training from scratch (``Scratch'') and from supervisedly trained models on Kinetics (``Supervised'').
For 3DRotNet, we compare to its model trained with RGB inputs, where 64f means 64 frames.
$^{\dag}$: AlexNet results are all pretrained on UCF101.
}
\label{tab:finetuneUH_usual}
\end{table}
\vspace{-4pt}

\setlength{\fboxrule}{1pt}%
\begin{table}[t]
\centering
\begin{tabular}{c|c|cc}
\hline
Networks & Algorithms & UCF & HMDB \\
\hline
\multirow{4}{*}{3DResNet} & Scratch & 60.0 & 27.0 \\
 & DPC~\cite{DPC} & 68.2 & 34.5 \\
 & VIE (ours) & \textbf{75.5} & \textbf{44.6} \\
 & \emph{Supervised} & \emph{84.8} & \emph{60.2} \\
\hline
\multirow{3}{*}{SlowFast} & Scratch & 70.0 & 37.0 \\
 & VIE (ours) & \textbf{78.9} & \textbf{50.1} \\
 & \emph{Supervised} & \emph{89.7} & \emph{70.4} \\
\hline
\multirow{3}{*}{ResNet18} & Scratch & 57.3 & 23.9 \\
 & VIE (ours) & 73.1 & 41.2 \\
 & \emph{Supervised} & \emph{83.5} & \emph{52.9} \\
\hline
\multicolumn{2}{c|}{VIE-TwoPathway-SF} & \textbf{80.4} & \textbf{52.5} \\
\hline
\end{tabular}
\caption{
Top-1 finetuning results on UCF101 and HMDB51 datasets using models pretrained on Kinetics \textbf{with} color-noise augmentation.
}
\label{tab:finetune_color}
\end{table}

\textbf{Transfer to static object categorization.}
To determine the extent to which the VIE procedure learns general visual representations, we also evaluate the learned representations for transfer to image categorization in ImageNet.
For models requiring multi-frame inputs, we generate a ``static video'' by tiling still images across multiple frames.
Results are shown in Table~\ref{tab:transKN}.
As they for the action recognition transfer, the two-pathway models are highest performing for this task as well. 
Interestingly, however, unlike for the case of action recognition, the multi-frame dynamic models are substantially worse than the single frame models on the ImageNet transfer task, and show a performance drop at the highest convolutional layers.
In fact, the transfer performance of the single-frame unsupervised model trained on Kinetics is actually \emph{better} than that of any model \emph{supervised} on Kinetics.
Taken together, these results strongly motivate the two-pathway architecture, as features that contribute to high performance on action recognition --- e.g. processing of dynamical patterns --- are not optimal for static-image performance. 
However, the relatively high performance of the static and two-pathway models shows that VIE can achieve useful generalization, even when train and test datasets are as widely divergent as Kinetics and ImageNet. 
\vspace{-4pt}

\section{Analysis} \label{sec:analysis}
\textbf{Benefit from long-range temporal structure.}
A key idea in \ShortName is to embed entire videos into the latent space, which is intended to leverage contextual information contained in the video. 
This may even have utility for videos with multiple scenes containing widely-diverging content (common for Kinetics), as the high-dimensional embedding might learn to situate such videos in the latent space so as to retain this structure.
As a preliminary test of the validity of this approach, we generated new training datasets by dividing each video into equal-length bins and then use these temporally-clipped datasets to train \ShortName.
Table~\ref{tab:ablation} show that the full-video model outperforms both 2- and 5-bin models, especially on ImageNet transfer learning performance, supporting the choice of embedding entire videos and also indicating that even better performance may be obtained using longer, more contextually complex videos.

\textbf{Benefit from more data.}
Although \ShortName achieves state-of-the-art unsupervised transfer performance on action recognition, and creates representations more useful for static object categorization than supervision on action recognition,
its learned representation is (unsurprisingly) worse on ImageNet categorization than its counterpart directly trained on the (much larger) ImageNet training set~\cite{zhuang2019local}.
To test whether \ShortName would benefit from more videos, we retrain \ShortName-Single with subsampled Kinetics (see Table~\ref{tab:ablation}).
Performance on ImageNet increases consistently and substantially without obvious saturation, indicating \ShortName's representation generalizability would benefit substantially if trained on a video dataset of the scale of ImageNet.

\begin{figure*}
\begin{center}
\includegraphics[width=.9\textwidth] {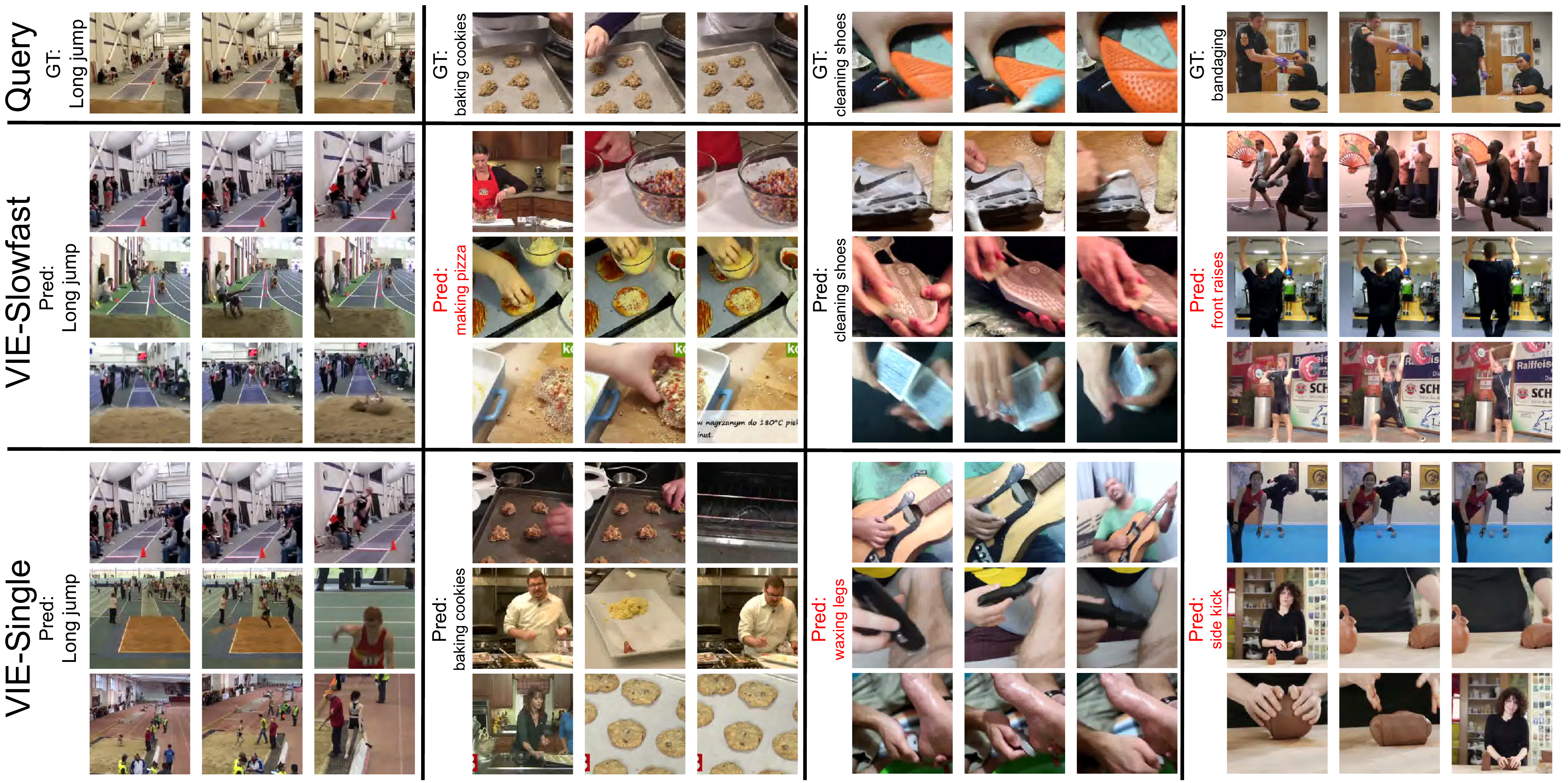}
\end{center}
\vspace{-4mm}
\caption{
Video retrieval results for \ShortName-Single and \ShortName-Slowfast models from Kinetics validation set.
GT=ground truth label, Pred=model prediction.
For each query video, three nearest training neighbors are shown.
Red font indicates an error.
}
\label{fig:retri}
\end{figure*}

\textbf{Video retrieval.}
We conduct a video retrieval experiment using distance in the embedding space.
Representative examples are shown in Figure~\ref{fig:retri}.
Oftentimes, qualitatively similar videos were retrieved, although some failure cases were also observed.
Moreover, \ShortName-Slowfast appears to extract context-free dynamic information, while \ShortName-Single is more biased by per-frame context,
further validating the idea that multi-frame models develop representations focusing on the dynamic features, while single-frame models better extract static information.
For example, in the ``cleaning shoes'' query, the two nearest \ShortName-Slowfast neighbors share a common dynamic (hand motions) with the query video, while hand and shoe position and the backgrounds all vary.
Meanwhile, \ShortName-Single only captures object semantics (the presence of the hand), lacking information about the movement that hand will make.
Retrieval failures likewise exemplify this result:
in the bandaging and baking cookies examples, \ShortName-Slowfast captures high-level motion patterns inaccessible to the static pathway.

\begin{table*}[t]
\centering
\begin{tabular}{c|ccc|ccc}
\hline
Dataset & \multicolumn{3}{|c|}{Kinetics} & \multicolumn{3}{|c}{ImageNet}\\
\hline
Layer & Conv3 & Conv4 & Conv5 & Conv3 & Conv4 & Conv5 \\
\hline\hline
\ShortName-Single & 23.84 & 38.25 & 44.41 & 25.02 & 40.49 & 42.33 \\
\hline
70\%-\ShortName-Single & 26.18 & 38.87 & 43.59 & 23.05 & 39.63 & 39.85 \\
30\%-\ShortName-Single & 25.54 & 37.49 & 40.72 & 23.33 & 38.49 & 36.23 \\
\hline
2bin-\ShortName-Single & 24.54 & 39.16 & 44.24 & 25.55 & 41.43 & 39.36 \\
5bin-\ShortName-Single & 25.17 & 38.73 & 43.33 & 23.90 & 40.46 & 37.83 \\
\hline
\end{tabular}
\caption{
Top-1 accuracy (in \%) of transfer learning to Kinetics and ImageNet from \ShortName-Single models trained using different amount of videos or with videos cut into different number of bins.
}
\label{tab:ablation}
\end{table*}

\section{Conclusion}
We have described the VIE method, an approach that combines multi-streamed video processing architectures with unsupervised deep embedding learning, and shown initial evidence that deep embeddings are promising for large-scale unsupervised video learning. 
In this work, our goal is not so much to illustrate the dominance of one particular novel loss function and architecture, but rather to provide a clear demonstration of the fact that a previously very challenging goal --- unsupervised learning on at-scale video tasks --- has become substantially more effective that it had previously been, due to a combination of very recent architectural and loss-function ideas. 
Together with recent results in static image categorization~\cite{zhuang2019local,wu2018unsupervised,CMC,caron2018deep}, our results suggest that the deep embedding approach is an increasingly viable framework for general unsupervised learning across many visual tasks. 

We've also found that performance across different types of visual tasks depends in an understandable fashion on how frame sampling is accomplished by the architecture, and that a two-pathway model with both static and dynamic pathways is relatively better than either alone.
This result is, interestingly, consistent with observations from neuroscience, where 
it has been shown that both ventral stream~\cite{yamins2014performance} and dorsal stream~\cite{goodale1992separate} brain pathways contribute to visual performance, with the former being more sensitive to static objects and the latter more sensitive to dynamic stimuli. 

However, there are a number of critical limitations in the current method that will need to be overcome in future work. Our choice of Local Aggregation for evaluation in this work should not taken as a statement that it is the only way to achieve useful unsupervised deep neural embeddings.
In particular, exploring the use of other very recent unsupervised learning losses, such as CMC~\cite{CMC} and DIM~\cite{DIM}, will also be of great interest.
We expect many of these methods would be compatible with the general VIE framework, and would likely be usefully complementary to LA. 

Another natural direction for improvement of the architecture is to investigate the use of recurrent neural network motifs~\cite{hochreiter1997long, nayebi2018task} and attentional mechanisms~\cite{vaswani2017attention}. 
It is likely that improved architectures of these or some other type could better take advantage of the rich medium-range temporal structure (e.g. $\sim$1s-1m in length) of natural video sequences.

Further, our current results are likely impacted by limitations in the Kinetics dataset, especially for harnessing the importance of dynamic processing, since even in the supervised case, single-frame performance is comparatively high. 
Seeking out and evaluating \ShortName on additional datasets will be critical --- perhaps most importantly, for applications involving large and previously uncurated video data where the potential impact of unsupervised learning is especially high. 
It will also be critical to test \ShortName in video task domains other than classification, including object tracking, dynamic 3D shape reconstruction and many others.


\noindent\textbf{Acknowledgments.}
This work was supported by the McDonnell Foundation (Understanding Human Cognition Award Grant No. 220020469), the Simons Foundation  (Collaboration on the Global Brain Grant No. 543061), the Sloan Foundation (Fellowship FG-2018-10963), the National Science Foundation (RI 1703161 and CAREER Award 1844724), and the National Institutes of Health (R01 MH069456), and hardware donation from the NVIDIA Corporation.

{\small
\bibliography{references}
\bibliographystyle{ieee_fullname}
}

\clearpage
\appendix
\title{Supplementary Material}
\date{}
\maketitle

\subsubsection*{LA Specific Parameters} \label{app:la}
For the LA-specific parameters, we use cluster size $m=8000$ for constructing close neighbors $\CloseNei$ and nearest neighbor size $k=512$ for constructing background neighbors $\BackNei$. These parameters depart somewhat from the optimal parameters found in ~\cite{zhuang2019local}, due to the substantial difference in size, and thus density in the embedding space, between the Kinetics training set (240K points) and the ImageNet dataset used in ~\cite{zhuang2019local} (1.2M points).

\subsubsection*{Network Implementation Details} \label{app:net}
For \ShortName-Single, we directly apply the ResNet-18 architecture and follow exactly the same preprocessing pipeline as described in the main text. 

For \ShortName-3DResNet, in order to be comparable to other works~\cite{Kim2018, Jing2018} which use a smaller input resolution for their networks, we correspondingly scale down our input image size.
More specifically, during training, we first resize the chosen frames so that their shortest edges are between $128$ and $160$px and then get $112\times 112$ images through random crops.
We then apply the same color noise and random horizontal flip to get the final inputs to the networks.
During testing, the frames are resized so that their shortest side is 128px, and then the center $112 \times 112$ crops are chosen as inputs.
Same as in~\cite{Kim2018, Jing2018}, the input clip contains 16 consecutive frames.

For \ShortName-TRN, we sample four consecutive half-second bins, and then one frame from each bin, using ResNet-18 as the shared 2D-CNN across multiple frames, with the outputs of the \textbf{Conv5} concatenated channel-wise and input into a fully-connected layer to generate the final embedding. 
This is a simplified version of the TRN, which runs faster and achieves only slightly lower supervised action-recognition performance than the full 8-frame TRN introduced in~\cite{TRN}.

For \ShortName-Slow and \ShortName-SlowFast, we follow ~\cite{slowfast} but modify it to use ResNet-18 rather than ResNet-50. The Slow model/pathway evenly samples one frame from every 16 to assemble a 4-frame input sequence, while the Fast pathway samples one frame from every 4 to assemble a 16-frame input sequence.

\subsubsection*{Single-frame Models with Multi-frame Inputs} \label{app:inp}
To control for the fact that multi-frame models received more total inputs than single-frame models, we also built models which, for any given multi-frame model, takes \ShortName-Single model, applies it to multiple frames using the same sampling strategy as for the multi-frame model, and then averages across the per-frame outputs before training the softmax classifier. These models are denoted with Input-Single. And their performance is shown in Table~\ref{tab:app_inp}.

\begin{table}[!ht]
\centering
\begin{tabular}{c|ccc}
\hline
Models & Conv3 & Conv4 & Conv5 \\
\hline\hline
TRN-Input-Single & 25.52 & 39.25 & 44.27 \\
Slow-Input-Single & 26.17 & 39.24 & 44.62\\
Sf-Input-Single & 25.72 & 39.38 & 44.29 \\
\hline
\end{tabular}
\renewcommand\thetable{S1} 
\caption{
Top-1 transfer learning accuracy (\%) on Kinetics for Input-Single models.
}
\label{tab:app_inp}
\end{table}

\subsubsection*{Reimplementation Details}
We reimplemented OPN~\cite{lee2017unsupervised}, RotNet~\cite{rotnet}, and 3DRotNet~\cite{Jing2018} methods and train them on Kinetics videos, as controls for \ShortName.
The implementation of OPN follows the procedure described in the paper as closely as possible, including input size, motion-related frame sampling, the use of frame-wise spatial jittering and channel dropping, and the learning rate schedule.
However, for a fair comparison, we use ResNet-18 as the OPN backbone. 
Our OPN implementation achieves approximately 40\% in the order prediction training task on Kinetics, similar to that reported in the original OPN paper, suggesting it is functioning as intended.
As for RotNet and 3DRotNet, we use ResNet-18 and 3DResNet-18 as the backbones respectively. The input resolution for 3DResNet-18 is set as $112 \times 112$, matching the input resolution of \ShortName-3DResNet. Other details follow the procedure described in the original papers.

\subsubsection*{Fine-tuning Implementation Details} \label{app:sml}
In testing for both preprocessing pipelines, each video is split into consecutive 16-frame clips and the outputs of all clips are averaged to get the final prediction. 
As for other parameters, the initial learning rate is 0.01 and the weight decay is 1e-4 for the training from scratch.
For finetuning, the initial learning rate is 0.0005 and the weight decay is 1e-5. 
The learning rate is dropped by 10 after validation performance saturates.
We report the results on the first split for both UCF101 and HMDB51, which should be close to the 3-split average result.

\end{document}